\documentclass{article}

\usepackage{microtype}
\usepackage{graphicx}
\usepackage{subfigure}
\usepackage{booktabs} % for professional tables
\usepackage{amsmath}

\usepackage{hyperref}

\usepackage[accepted]{icml2019}

\icmltitlerunning{Improving Missing Data Imputation with Deep Generative Models}

\begin{document}

\twocolumn[
\icmltitle{Improving Missing Data Imputation with Deep Generative Models}

% It is OKAY to include author information, even for blind
% submissions: the style file will automatically remove it for you
% unless you've provided the [accepted] option to the icml2019
% package.

% List of affiliations: The first argument should be a (short)
% identifier you will use later to specify author affiliations
% Academic affiliations should list Department, University, City, Region, Country
% Industry affiliations should list Company, City, Region, Country

% You can specify symbols, otherwise they are numbered in order.
% Ideally, you should not use this facility. Affiliations will be numbered
% in order of appearance and this is the preferred way.
\icmlsetsymbol{equal}{*}

\begin{icmlauthorlist}
\icmlauthor{Ramiro D. Camino}{unilu}
\icmlauthor{Christian A. Hammerschmidt}{unilu}
\icmlauthor{Radu State}{unilu}
\end{icmlauthorlist}

\icmlaffiliation{unilu}{University of Luxembourg, Luxembourg}

\icmlcorrespondingauthor{Ramiro D. Camino}{ramiro.camino@uni.lu}
\icmlcorrespondingauthor{Christian A. Hammerschmidt}{chrishammerschmidt@posteo.de}
\icmlcorrespondingauthor{Radu State}{radu.state@uni.lu}

% You may provide any keywords that you
% find helpful for describing your paper; these are used to populate
% the "keywords" metadata in the PDF but will not be shown in the document
\icmlkeywords{Machine Learning, Deep Learning, Generative Models, Missing Data Imputation, ICML}

\vskip 0.3in
]

% this must go after the closing bracket ] following \twocolumn[ ...

% This command actually creates the footnote in the first column
% listing the affiliations and the copyright notice.
% The command takes one argument, which is text to display at the start of the footnote.
% The \icmlEqualContribution command is standard text for equal contribution.
% Remove it (just {}) if you do not need this facility.

\printAffiliationsAndNotice{}  % leave blank if no need to mention equal contribution
%\printAffiliationsAndNotice{\icmlEqualContribution} % otherwise use the standard text.

\begin{abstract}
% Note from paper example:
% Keep your abstract brief and self-contained, one paragraph and roughly 4-6 sentences.
% Gross violations will require correction at the camera-ready phase.

% Purpose and Motivation
Datasets with missing values are very common on industry applications, and they can have a negative impact on machine learning models.
Recent studies introduced solutions to the problem of imputing missing values based on deep generative models.
% Problem
Previous experiments with Generative Adversarial Networks and Variational Autoencoders showed interesting results in this domain, but it is not clear which method is preferable for different use cases.
% Proposal
The goal of this work is twofold: we present a comparison between missing data imputation solutions based on deep generative models, and we propose improvements over those methodologies.
% Methodology
We run our experiments using known real life datasets with different characteristics, removing values at random and reconstructing them with several imputation techniques.
% Results
Our results show that the presence or absence of categorical variables can alter the selection of the best model, and that some models are more stable than others after similar runs with different random number generator seeds.
% Conclusion
\end{abstract}
\section{Introduction}

% data science
Analyzing data is a core component of scientific research across many domains. 
% reproducible science
Over the recent years, awareness for the need of transparent and reproducible work increased. 
This includes all steps that involve preparing and pre-processing the data.
An increasingly common pre-processing step is the imputation of missing values~\cite{hayati_rezvan_rise_2015}.
% why missing data is an issue
Data with missing values can decrease model quality and even lead to wrong insights~\cite{lall_how_2016} by introducing biases. 
Likewise, dropping samples with missing values can cause larger errors because of the scarce amount of remaining data.
% define missing data imputation
One solution is performing data imputation, which consists in replacing missing values with substitutes.
% types of missing data is important
However, the reason behind the missingness needs to be clarified before imputing \cite{rubin1976inference}.
% example of bad data imputation
For example, \cite{lall_how_2016}  show that in political sciences, results often do not hold up if missing data is imputed improperly.
% mcar
Values are considered to be missing completely at random (MCAR) when the probability that they are missing is independent both on the value and on other observable values of the data.
% mar
The case when data is missing at random (MAR) happens when the missing probability can be estimated from variables where the value is present.
% mnar
Finally, the case when missing data is neither MCAR nor MAR, is defined as missing not at random (MNAR).
This means that the reason for a value to be missing, can depend on other variables, but also on the value that is missing.
% type used on this paper
In order to be able to compare the desired models in the same framework, we assume for the this study that missing values are MCAR.

% goal of this paper
In this paper, we survey the quickly evolving state-of-the-art of deep generative models for tabular data and missing value imputation.
% backpropagation imputation
We propose using a backpropagation technique to correct the imputed values iteratively.
% variable split architecture
We also adapt the analyzed models by changing the inputs and outputs of the architectures, taking into account the size and the type of each sample variable.
% experiments
Furthermore, we compare the imputation power of every technique using real life datasets.

% These layers improve performance across all models.
% Beyond surveying the quickly evolving state-of-the-art, we also propose extending existing generative architectures with multi-input and multi-output layers to handle with multiple variables.
\section{Related Work}

Work related to our approach falls into three groups.
The first one consists of state-of-the-art imputation algorithms.
The second group is composed by generative models based in neural networks, and in particular, networks focusing on generating tabular data and handling issues related to categorical variables, rather than generating one high-dimensional image or text variable.
Lastly, the third group is constituted by methodologies using deep generative models for imputation in the domain of tabular data.

Within the field of missing value imputation, traditional methods can be classified into discriminative and generative imputation models.
Examples of discriminative models with state-of-the-art performance are MICE~\cite{buuren2010mice}, MissForest \cite{stekhoven2011missforest}, and Matrix Completion \cite{mazumder2010spectral}.
Autoencoders \cite{gondara2017multiple} and Expectation Maximization \cite{garcia2010pattern} are instances of generative models.
Key distinguishing factors of these methods are limitations coming from necessary assumptions about the nature and distribution of the data and the ability to learn from samples with missing data (rather than only learning from complete data samples).

% intro to deep generative models
% TODO: cite examples of use cases?
Deep generative models like Variational Autoencoders (VAE) \cite{kingma2013auto} and Generative Adversarial Networks (GAN) \cite{goodfellow2014generative} proved to be very powerful in the domain of computer vision \cite{brock_large_2018}, speech recognition and natural language processing \cite{DBLP:journals/corr/JainZS17, DBLP:journals/corr/LinLHZS17}.
% extend to other domains
It is expected for the scientific community to extend the application of these models to other areas.
% medGAN
The authors of medGAN \cite{choi_generating_2017} applied GANs to generate synthetic health care patient records represented by numerical and binary features.
% multi-categorical GAN
The multi-categorical GANs \cite{camino2018generating} extended medGAN and other architectures by splitting the outputs of the networks into parallel layers depending on the size of categorical variables, and used gumbel-softmax activations \cite{jang_categorical_2016,maddison_concrete_2016} to handle discrete distributions.
% GAN for PNR
\cite{mottini2018airline} proposed a GAN based architecture to generate synthetic passenger name records, dealing with missing values and a mix of categorical and numerical variables.
% TGAN
Tabular GAN (TGAN) \cite{xu2018synthesizing} presented a method to generate data composed by numerical and categorical variables, where the generator outputs variable values in an ordered sequence using a recurrent neural network architecture.

% image completion
There are numerous studies related to image completion with deep generative models like \cite{vincent2008extracting}, that uses denoising autoencoders for image imputing.
% sentence completion
In the domain of natural language processing, \cite{bowman2015generating} presented a VAE model with a recurrent architecture for sentence generation and imputation.
% extension to missing value imputation
This use case was also translated to the topic of missing value imputation on tabular data.
% GAIN
Generative Adversarial Imputation Nets (GAIN) \cite{yoon2018gain} adapted the GAN architecture to this problem and proved to be more efficient than state-of-the-art imputation methods.
% HI-VAE
The Heterogeneous-Incomplete VAE (HI-VAE) \cite{nazabal2018handling} proposed an imputation technique for tabular data based on VAE, and also compared their results to state-of-the-art imputation methodologies.

\section{Approach}

\subsection{Problem Definition}

% a dataset is composed by samples
We define a tabular dataset $X$ as a collection of samples $\{x_1, \ldots, x_n\}$.
% a sample is composed by variable values
Each sample $x_i$ is a collection of values $\{v_{i1}, \ldots, v_{im}\}$ for the variables $\{V_1, \ldots, V_m\}$, where each variable $V_j$ has a type $t_j \in \{$numerical, categorical$\}$.
% numerical variable
The value of a numerical variable $V_j$ is represented only by one real valued feature, hence we define the size of a numerical variable as $s_j = 1$.
% categorical variables
A categorical variable $V_j$ can take one of $s_j > 1$ possible categories.
We one-hot encode each value $v_{ij}$ into $s_j$ binary features, by turning on the feature corresponding to the selected category and turning off the remaining $s_j - 1$ features.
% number of features
Combining the amount of features of each variable, we define the amount of features (or size) of every sample as $s = \sum_{j=1}^m s_j$.

% missing values
We can now define a dataset with missing values $\bar{X}$ as a copy of a dataset $X$ where one or more variable values from one or more samples were dropped.
% imputation
An imputation algorithm or model $\mathcal{M}$ takes a dataset with missing values $\bar{X}$ and outputs a dataset $\hat{X}$ by filling the missing values of $\bar{X}$.
The goal of these algorithms is to minimize the difference between the original dataset $X$ and its reconstructed version $\hat{X}$.

% mask
In addition, we define a the mask $M$ of $\bar{X}$ that will be used in the following method definitions.
We represent $M$ as a matrix $\in \{0, 1\}^{n \times s}$, where each position $m_{ik}$ contains a 1 if the feature on $\bar{x}_{ik}$ is present or a 0 if it is missing.
% mini-batches
Also for the training of deep learning models, we note subset of sample indices $B \subset \{1, \ldots, n\}$ as a mini-batch.

\subsection{Gumbel-Softmax}

The output of a neural network can be transformed into a multinominal distribution by passing it through a softmax layer.
However, sampling from this distribution is not a differentiable operation, which blocks the backpropagation during the training of generative models for discrete samples. 
The Gumbel-Softmax \cite{jang_categorical_2016} and the Concrete-Distribution \cite{maddison_concrete_2016} were simultaneously proposed to tackle this problem in the domain of variational autoencoders (VAE) \cite{kingma_auto-encoding_2013}.
Later \cite{kusner_gans_2016} adapted the technique to GANs for sequences of discrete elements.

For i.i.d samples  $g_1, \ldots, g_d$  drawn from $Gumbel(0, 1) = -\log(-\log(u_i))$ with $u_i \sim U(0, 1)$, 
the gumbel-softmax generates sample vectors $b \in \left[0, 1\right]^d$ based on inputs $a \in R^d$ (that can be the output of previous layers) and a temperature hyperparameter $\tau \in (0, \infty)$ by the formula:

\begin{equation*}
\begin{array}{rl}
b_i = \dfrac{exp((\log(a_i) + g_i) / \tau)}{\sum_{j=1}^d exp((\log(a_j) + g_j) / \tau)} & i=1, \ldots, d 
\end{array}
\end{equation*}

\subsection{GAIN}

% based on GAN
GAIN \cite{yoon2018gain} is composed of a modified generator and discriminator network.
Compared to the original GAN architecture~\cite{goodfellow2014generative}, the differences are as follows:
% generator
The generator $G$ takes as input some data $\bar{X}$, a suitable mask $M$, and a source of noise. 
In practice, the noise is inserted inside $\bar{X}$ in the positions of the missing values.
The model then returns an imputation $\hat{X}$.\newline
% discriminator
The discriminator $D$ of the GAN architecture receives $\hat{X}$, and instead of trying to determine if each sample from the input is either real or fake, the model tries guess for every sample if each variable value is either original or imputed.
% prediction
In other words, the discriminator needs to be trained to maximize the probability of predicting the mask $M$.
Hence, we note the output of network $D$ as $\hat{M}$.
% discriminator loss
The discriminator loss only takes into account the binary cross entropy of the positions corresponding to missing values:

\begin{small}
\begin{equation*}
\mathcal{L}_D(M_i, \hat{M}_i) =
- \sum_{k=1}^{s} m_{ik} \log \hat{m}_{ik} + (1 - m_{ik}) \log (1 - \hat{m}_{ik})
\end{equation*}
\end{small}

\noindent for every sample $i$. Then, for a mini-batch $B$, the discriminator $D$ with parameters $\theta_D$ is trained to optimize:

\begin{equation*}
\min_{\theta_D} \sum_{i \in B} \mathcal{L}_D(M_i, \hat{M}_i)
\end{equation*}

% generator loss
Given that $\hat{M} = D(\hat{X})$ and $\hat{X} = G(\bar{X}, M)$, the generator $G$ with parameters $\theta_G$ is trained to maximize the probability of fooling the discriminator predictions of $M$:

\begin{equation*}
\mathcal{L}_G(M_i, \hat{M}_i) = - \sum_{k=1}^{s} (1 - m_{ik}) \log \hat{m}_{ik}
\end{equation*}

\begin{equation*}
\min_{\theta_G} \sum_{i \in B} \mathcal{L}_G(M_i, \hat{M}_i)
+ \alpha \mathcal{L}_{rec}(\bar{X}_i, \hat{X}_i, M_i)
\end{equation*}

% reconstruction loss term
\noindent where $\mathcal{L}_{rec}(\bar{X}_i, \hat{X}_i, M_i)$ is a special reconstruction loss term weighted by the hyperparameter $\alpha$.
% reconstruction loss term by variable and mask
This loss function is separated by variable and masked in order to calculate only the reconstruction error for the non-missing values:

\begin{equation}
\label{eq:rec-error}
\mathcal{L}_{rec}(\bar{X}_i, \hat{X}_i, M_i) =
\sum_{j=1}^m \sum_{q=1}^{s_j} m_{ik} rec_j(\bar{x}_{ik}, \hat{x}_{ik})
\end{equation}

\noindent where the feature index $k$ is calculated as:

\begin{equation*}
k = \sum_{j'=0}^{j-1} s_{j'} + q - 1
\end{equation*}

\noindent which is the index for the first feature of $v_j$ plus the offset for the categorical value $q$ (or no offset for numerical variables).
The reconstruction error for each feature of a particular variable $v_j$ depends on the variable type $t_j$:

\begin{equation*}
rec_j(\bar{x}_{ik}, \hat{x}_{ik}) = \left\{\begin{array}{ll}
\left( \hat{x}_{ik} - \bar{x}_{ik} \right)^2
& \text{if } t_j \text{ = numerical} \\
- \bar{x}_{ik} \log \hat{x}_{ik}
& \text{if } t_j \text{ = categorical}
\end{array}\right.
\end{equation*}

\subsection{VAE Imputation}

% original VAE
We train a traditional VAE \cite{kingma2013auto} using $\bar{X}_{train}$ as input, but inserting random noise on the positions with missing values.
% reconstruction loss mask
We use the masked the reconstruction loss defined in Equation \ref{eq:rec-error} to calculate the reconstruction error between $\hat{X}_{train}$ and $X_{train}$ only on the non-missing values.
% imputation
We use the trained model to obtain $\hat{X}_{test}$ from $\bar{X}_{test}$ and use the reconstructed missing values for imputation.
% maybe add the equations?

\subsection{VAE Iterative Imputation}

% extends VAE imputation
The method from \cite{mccoy2018variational} is an extension of the VAE imputation, related to the iterative PCA imputation algorithm \cite{dray2015principal}.
% differences with VAE imputation
The training stage remains the same, but the imputation procedure is corrected iteratively.
% stopping criteria
The reconstruction error of the non-missing values is measured after each iteration, and if the error is small enough or too many iterations have been calculated, the algorithm stops.
% reusing noise
If the procedure must continue, $\bar{X}_{test}$ is run through the model again, but instead of inserting random noise on the positions with missing values, the reconstructed missing values from the previous iteration are used.
% using the reconstruction error of missing values
Note that it makes no sense to use the reconstruction error of the missing values as a stopping criteria, because in a real case scenario, the missing values are supposed to be unknown.
% algorithm reference
A pseudocode for this method is presented in Algorithm \ref{alg:vae+it}.

\begin{algorithm}[tb]
   \caption{VAE Iterative Imputation}
   \label{alg:vae+it}
\begin{algorithmic}
   \STATE {\bfseries Input:} $\bar{X}_{train}$ data with missing values for training
   \STATE {\bfseries Input:} $M_{train}$ mask for $\bar{X}_{train}$
   \STATE {\bfseries Input:} $\bar{X}_{test}$ data with missing values to impute
   \STATE {\bfseries Input:} $M_{test}$ mask for $\bar{X}_{test}$
   \STATE {\bfseries Input:} $i_{max}$ maximum number of iterations
   \STATE {\bfseries Input:} $e_{min}$ minimum error
   \STATE {\bfseries Output:} $\hat{X}_{test}$ imputation of $\bar{X}_{test}$
   \STATE $VAE = train(\bar{X}_{train}, M_{train})$
   \STATE $\epsilon \sim\mathcal{N}(0, 1)^{sizeof(M_{test})}$
   \STATE $\hat{X}_{test} = \bar{X}_{test} * M_{test} + \epsilon  * (1 - M_{test})$
   \STATE $i = 0$
   \REPEAT
   \STATE $\hat{X}_{test} = VAE(\hat{X}_{test})$
   \STATE $e = \mathcal{L}_{rec}(\bar{X}_{test}, \hat{X}_{test}, M_{test})$
   \STATE $\hat{X}_{test} = \bar{X}_{test} * M_{test} + \hat{X}_{test}  * (1 - M_{test})$
   \STATE $i++$
   \UNTIL $e \leq e_{min}$ \OR $i \geq i_{max}$
\end{algorithmic}
\end{algorithm}

\subsection{VAE Backpropagation Imputation}

% modifies VAE iterative imputation
The thrid VAE technique is an altered form of the iterative imputation: on each iteration, the reconstruction error of the non-missing values is backpropagated to the input.
% modify the noise
Instead of updating the weights of the model like in a traditional training, the noise itself is updated, keeping the model and the non-missing values untouched.
% stopping criteria
The stopping criteria remains the same: the algorithm iterates until the error is sufficiently small or the number of iterations is large enough.
% based on
This technique is inspired by \cite{schlegl2017unsupervised}, where the authors detect anomalies on images by searching through the latent space used as input for the generator.
% algorithm reference
A pseudocode for this method is presented in Algorithm \ref{alg:vae+bp}, highlighting the differences with Algorithm \ref{alg:vae+it}.

\begin{algorithm}[tb]
   \caption{VAE Iterative Backpropagation Imputation}
   \label{alg:vae+bp}
\begin{algorithmic}
   \STATE {\bfseries Input:} $\bar{X}_{train}$ data with missing values for training
   \STATE {\bfseries Input:} $M_{train}$ mask for $\bar{X}_{train}$
   \STATE {\bfseries Input:} $\bar{X}_{test}$ data with missing values to impute
   \STATE {\bfseries Input:} $M_{test}$ mask for $\bar{X}_{test}$
   \STATE {\bfseries Input:} $i_{max}$ maximum number of iterations
   \STATE {\bfseries Input:} $e_{min}$ minimum error
   \STATE {\bfseries Output:} $\hat{X}_{test}$ imputation of $\bar{X}_{test}$
   \STATE $VAE = train(\bar{X}_{train}, M_{train})$
   \STATE $\epsilon \sim\mathcal{N}(0, 1)^{sizeof(M_{test})}$
   \STATE $\hat{X}_{test} = \bar{X}_{test} * M_{test} + \epsilon  * (1 - M_{test})$
   \STATE $i = 0$
   \REPEAT
   \STATE $\hat{X}_{test} = VAE(\hat{X}_{test})$
   \STATE $e = \mathcal{L}_{rec}(\bar{X}_{test}, \hat{X}_{test}, M_{test})$
   \STATE {\color{red} $grads = backprop(e)$}
   \STATE {\color{red} $\hat{X}_{test} = optim(\hat{X}_{test}, grads)$}
   \STATE $\hat{X}_{test} = \bar{X}_{test} * M_{test} + \hat{X}_{test}  * (1 - M_{test})$
   \STATE $i++$
   \UNTIL $e \leq e_{min}$ \OR $i \geq i_{max}$
\end{algorithmic}
\end{algorithm}

\subsection{Variable Splitting}

\begin{figure}[ht]
\vskip 0.2in
\begin{center}
\includegraphics[width=\columnwidth]{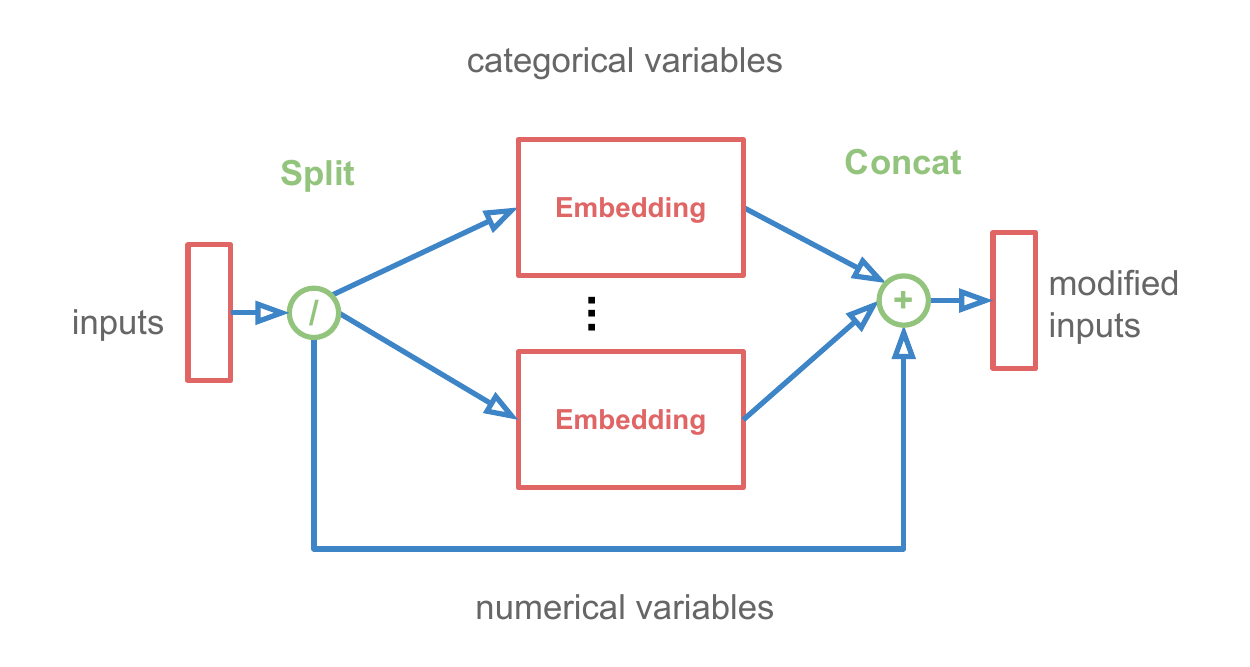}
\caption{Architecture for splitting input variables.}
\label{fig:vs-inputs}
\end{center}
\vskip -0.2in
\end{figure}

\begin{figure}[ht]
\vskip 0.2in
\begin{center}
\includegraphics[width=\columnwidth]{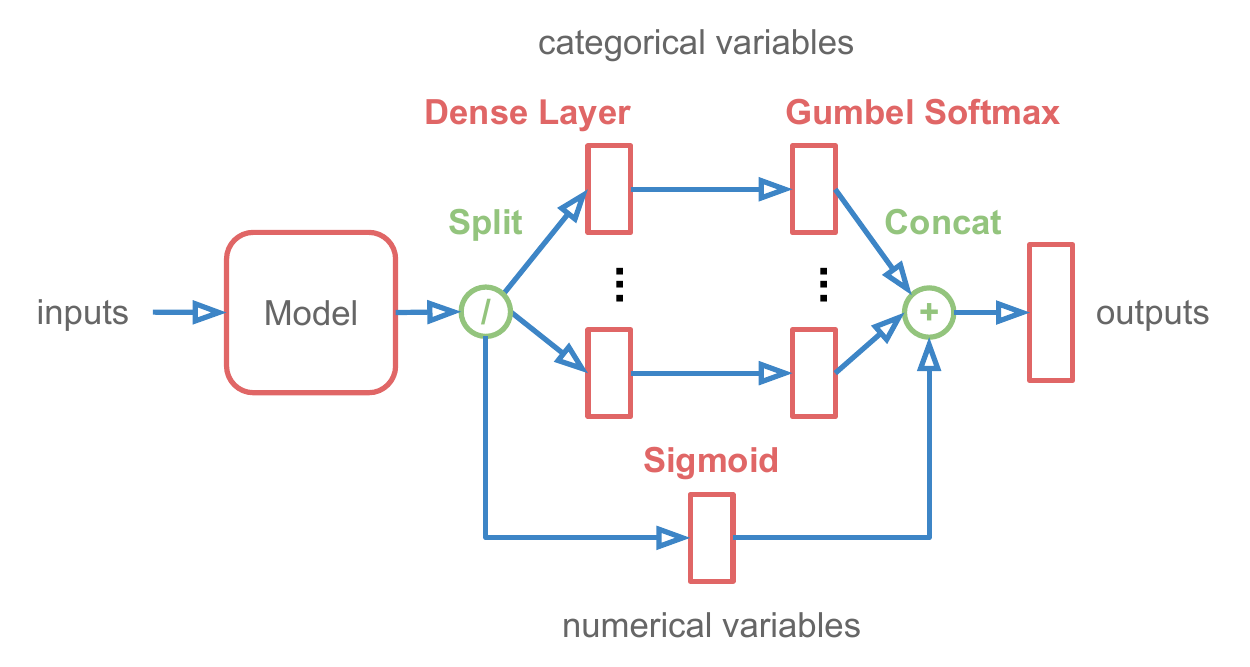}
\caption{Architecture for splitting output variables.}
\label{fig:vs-outputs}
\end{center}
\vskip -0.2in
\end{figure}

% based on multi-output GANs
The authors of multi-output GANs \cite{camino2018generating} introduced the notion of taking into account the structure of categorical variables from tabular data to modify the network architecture.
% parallel dense layers
A multi-output GAN contains one dense layer per variable connected in parallel to the output of the generator.
% dense layer sizes
Each dense layer transforms the output to the size of the corresponding variable.
% activations
A gumbel-softmax activation \cite{jang_categorical_2016} (or concrete distribution \cite{maddison_concrete_2016}) is used after each parallel dense layer.
% contatenation
All the outputs are concatenated back into a complete synthetic sample.
% autoencoders
The work on \cite{camino2018generating} also extended other GAN architectures involving autoencoders \cite{junbo_adversarially_2017,choi_generating_2017}, where the decoder is the altered model with multiple outputs, and the generator returns continuous latent codes for the autoencoders.
In this study we extend multi-output architecture in two ways:

\begin{itemize}
    % mixing numerical variables
    \item We work with tabular data of mixed numerical and categorical variables.
    In order to do so, during the splitting of numerical variables, we do not apply size transformations with dense layers, but we apply sigmoid activations before concatenating the output. The changes can be observed on Figure \ref{fig:vs-outputs}.
    This extends \cite{camino2018generating} to settings with mixed numerical and categorical data.
    % multiple embeddings
    \item We transfer the idea to the inputs too, by using one embedding layer per categorical variable with the corresponding size of each variable.The outputs of each parallel embedding layer are concatenated back to build the altered inputs. Numerical variables are concatenated directly with the outputs of the embeddings. Related architectures and applications can be found in \cite{weston2011wsabie, de2015artificial}. A representation can be seen on Figure \ref{fig:vs-inputs}.
\end{itemize}

% GAIN+vs
We propose to modify the GAIN architecture by adding the multiple-inputs and multiple-outputs both to the generator and the discriminator.
% VAE+vs
For the VAE, we propose to use multi-input on the encoder and multi-output on the decoder.
\section{Experiments}

\begin{table*}[ht]
\caption{Dataset properties.}
\label{tab:datasets}
\vskip 0.15in
\begin{center}
\begin{small}
\begin{sc}
\begin{tabular}{lccccc}
\toprule
Name & Samples & Features & Variables & Numerical & Categorical \\
\midrule
breast cancer & 569 & 30 & 30 & 30 & 0 \\
default credit card & 30000 & 93 & 23 & 14 & 9 \\
letter recognition & 20000 & 16 & 16 & 16 & 0 \\
online news popularity & 39644 & 60 & 47 & 44 & 3 \\
spambase & 4601 & 57 & 57 & 57 & 0 \\
\bottomrule
\end{tabular}
\end{sc}
\end{small}
\end{center}
\vskip -0.1in
\end{table*}

\subsection{Data and Pre-Processing}

% Data
We evaluate the presented imputation methods using five real life datasets from the UCI repository \cite{Dua:2017}.
They are composed of different number of samples and variables, both categorical and numerical.
Originally, no dataset contains missing values.
A summary of their properties is defined in Table \ref{tab:datasets}.

% Encoding and Scaling
For all datasets, categorical variables are one-hot encoded, and each numerical variable $V_j$ is scaled to fit inside the interval $[0, 1]$ according to the equation:

\[ scaled(V_j) = \frac{V_j - \min(V_j)}{\max(V_j) - \min(V_j)} \]

\noindent where $\max(V_j)$ and $\min(V_j)$ are the maximum and minimum values for the variable $V_j$ from all samples that can be measured across the entire dataset.

\subsection{Running the Experiments}

% Code
All the experiments are implemented using Python 2.7.15, PyTorch 0.4.0 \cite{paszke2017automatic} and scikit-learn 0.19.1 \cite{scikit-learn}.
The code is available in the supplementary material.
After the review process, we plan to port the implementation to Python 3 and release it to the public.

% Cross Validation
All models are trained with 90\% of the original data and the other 10\% is used to evaluate the performance of the imputation methods.
% Missing Value Generation
We generate different settings of missing value probabilities $p \in \{0.2, 0.5, 0.8\}$.
Given a missing probability $p$, each variable of each sample is dropped if $r < p$, for a random number $r \sim \mathcal{U}[0, 1]$.
The value of a dropped variable is replaced by random noise $\mathcal{N}(0, 1)$.

% Hyperparameters
Each method has different amount of hyperparameters that need to be defined.
% GAIN hyperparameters
For GAIN, we use the same configuration defined in \cite{yoon2018gain}, and we only experiment with different values for the weight of the reconstruction loss as the authors did originally.
Across all our experiments we get the best results with a value of 10.
% VAE hyperparameters: random search
We perform a random search for the hyperparameters needed by the VAE based methods.
% VAE hyperparameters: number and size of layers
Both the encoder's and the decoder's hidden layers (beside the special input and output layers) are implemented as a collection of fully connected layers with ReLU activations.
We experiment with one hidden layer with 50\% of the input size, two layers with 100\% and 50\% of the input size respectively, and no hidden layers at all.
% VAE hyperparameters: latent space size
For the size of the latent space, we tried with 10\%, 50\% and 100\% of the input size.
% VAE hyperparameters: batch size and learning rate
Other learning hyperparameters include the batch size, for which we explored the values $\{2^6, 2^7, 2^8, 2^9, 2^{10}\}$, and two different learning rates of $1e^{-3}$ and $1e^{-5}$.
% Iterative imputation hyperparameters
% Backpropagation iterative imputation hyperparameters 
Both of the iterative imputation also need to define the maximum number of iterations and the minimum acceptable error to stop, for which we fix 10000 iterations and $1e^{-4}$.
% Variable split hyperparameters
Finally, the variable split adapted architectures use gumbel-softmax activations, and we experiment with the temperature or $\tau$ hyperparameter using the values $\{0.1, 0.5, 1, 2, 10\}$.

% Seeds
Furthermore, every experiment is repeated at least three times using different seeds on the random number generator.
% Metric
For each collection of experiments that differ only on the seed, we calculate the mean and the standard deviation of the Root Mean Squared Error (RMSE) between the (normalized) imputed and the original test data values.

\subsection{Results}

\begin{figure*}[ht]
\minipage{0.32\textwidth}
  \includegraphics[width=\linewidth]{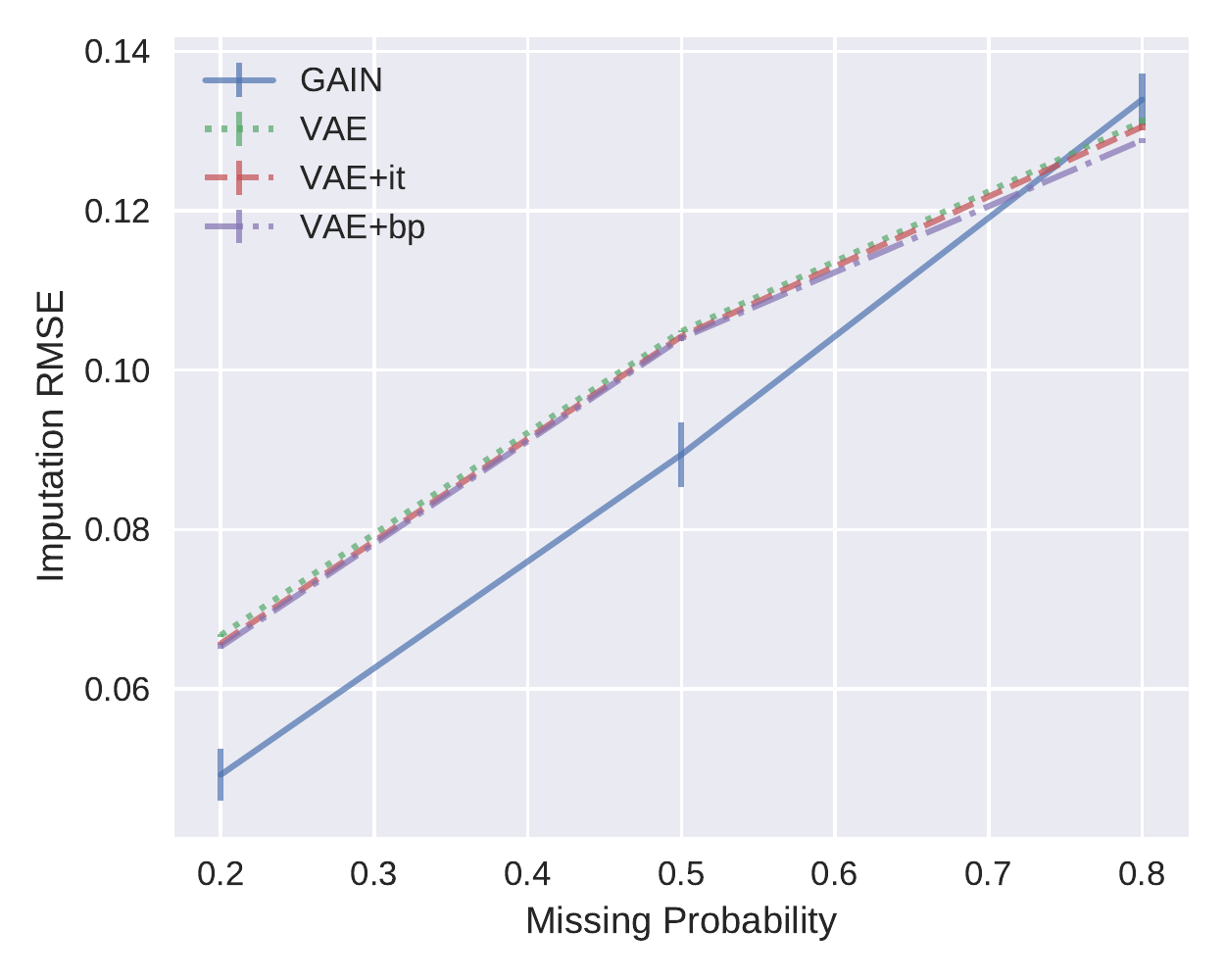}
\endminipage\hfill
\minipage{0.32\textwidth}
  \includegraphics[width=\linewidth]{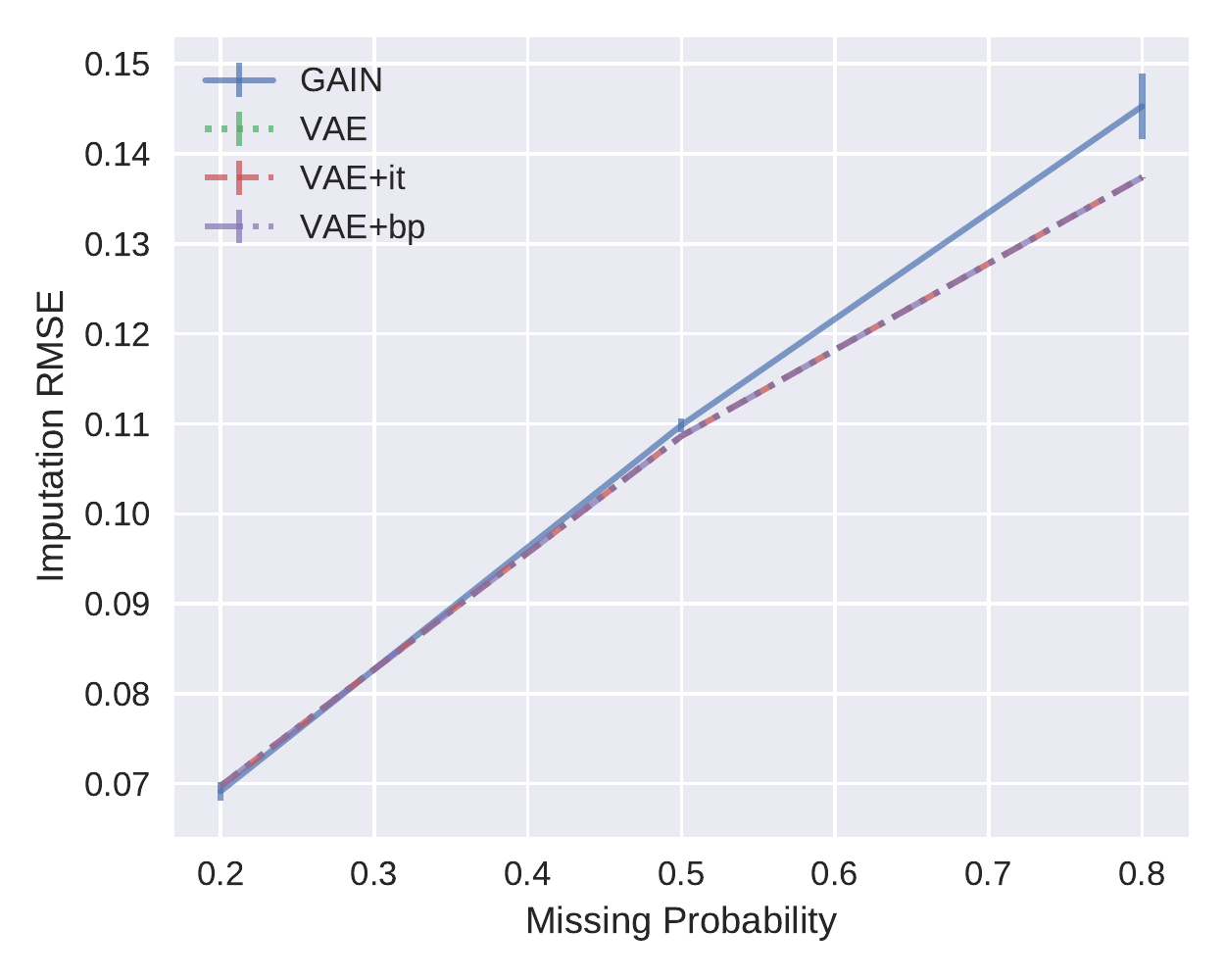}
\endminipage\hfill
\minipage{0.32\textwidth}%
  \includegraphics[width=\linewidth]{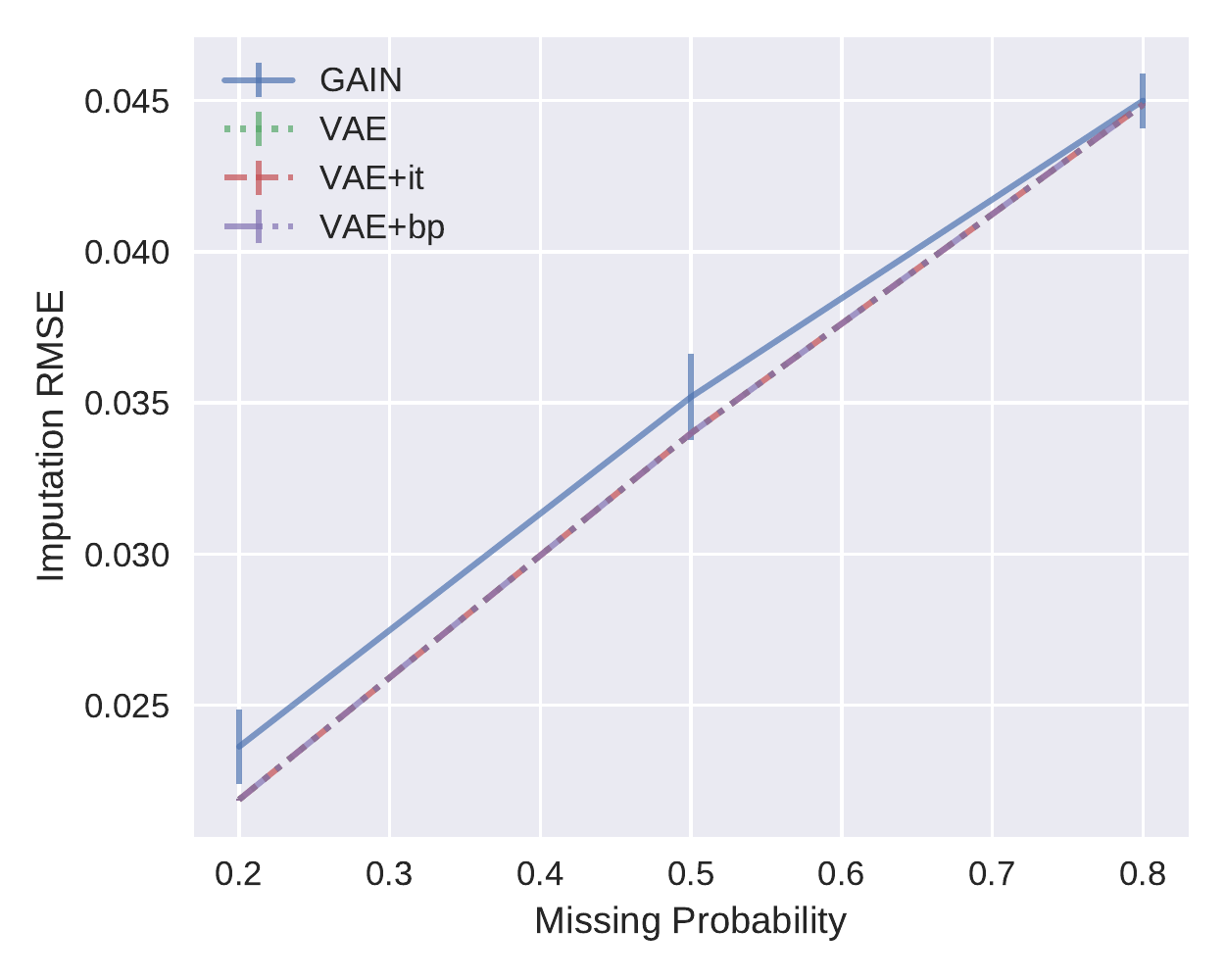}
\endminipage
\caption{Imputation RMSE by increasing proportion of missing values for the Breast Cancer, Letter Recognition and Spambase datasets.}
\label{fig:numerical}
\end{figure*}

\begin{figure*}[ht]
\minipage{0.44\textwidth}
  \includegraphics[width=\linewidth]{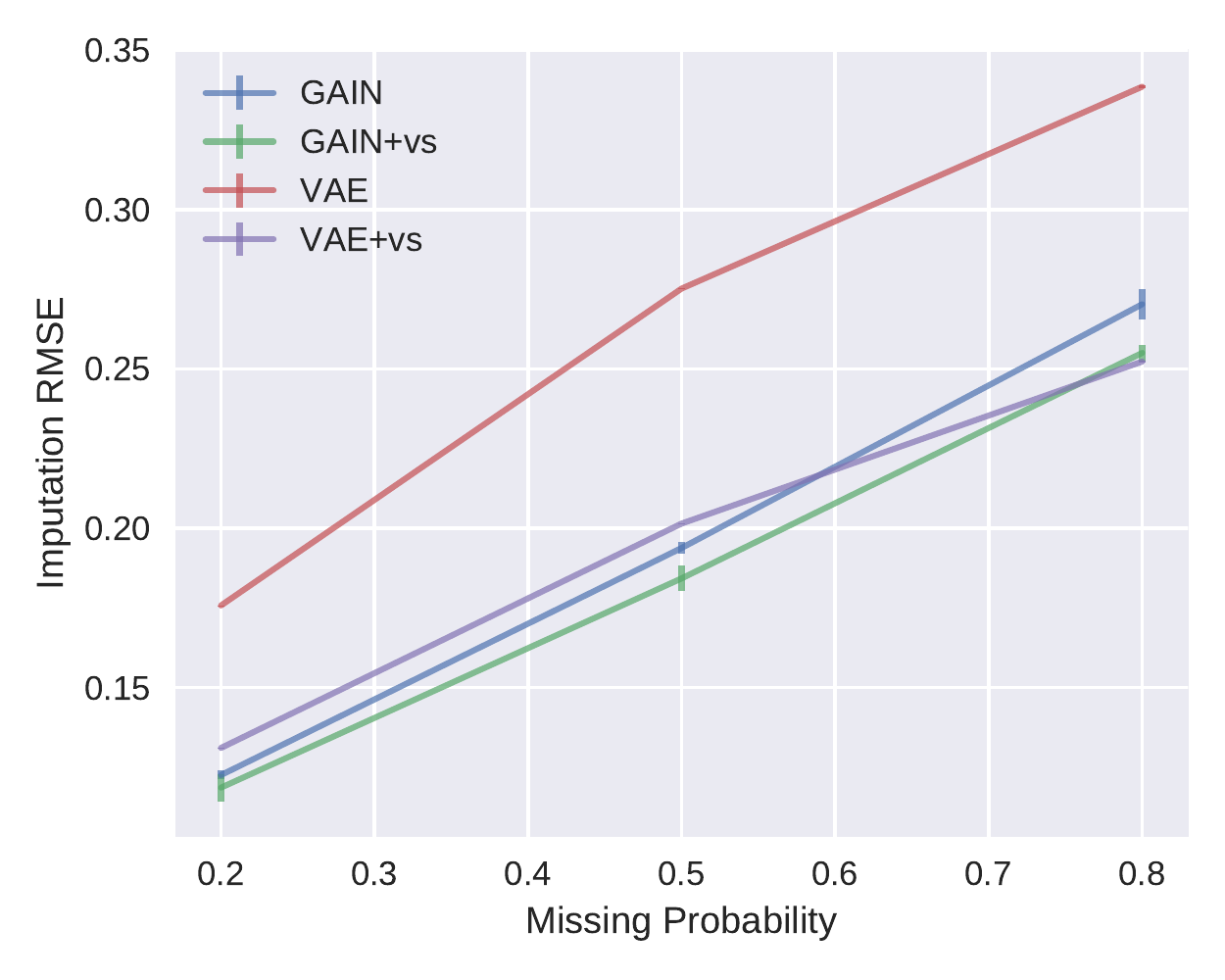}
\endminipage\hfill
\minipage{0.44\textwidth}%
  \includegraphics[width=\linewidth]{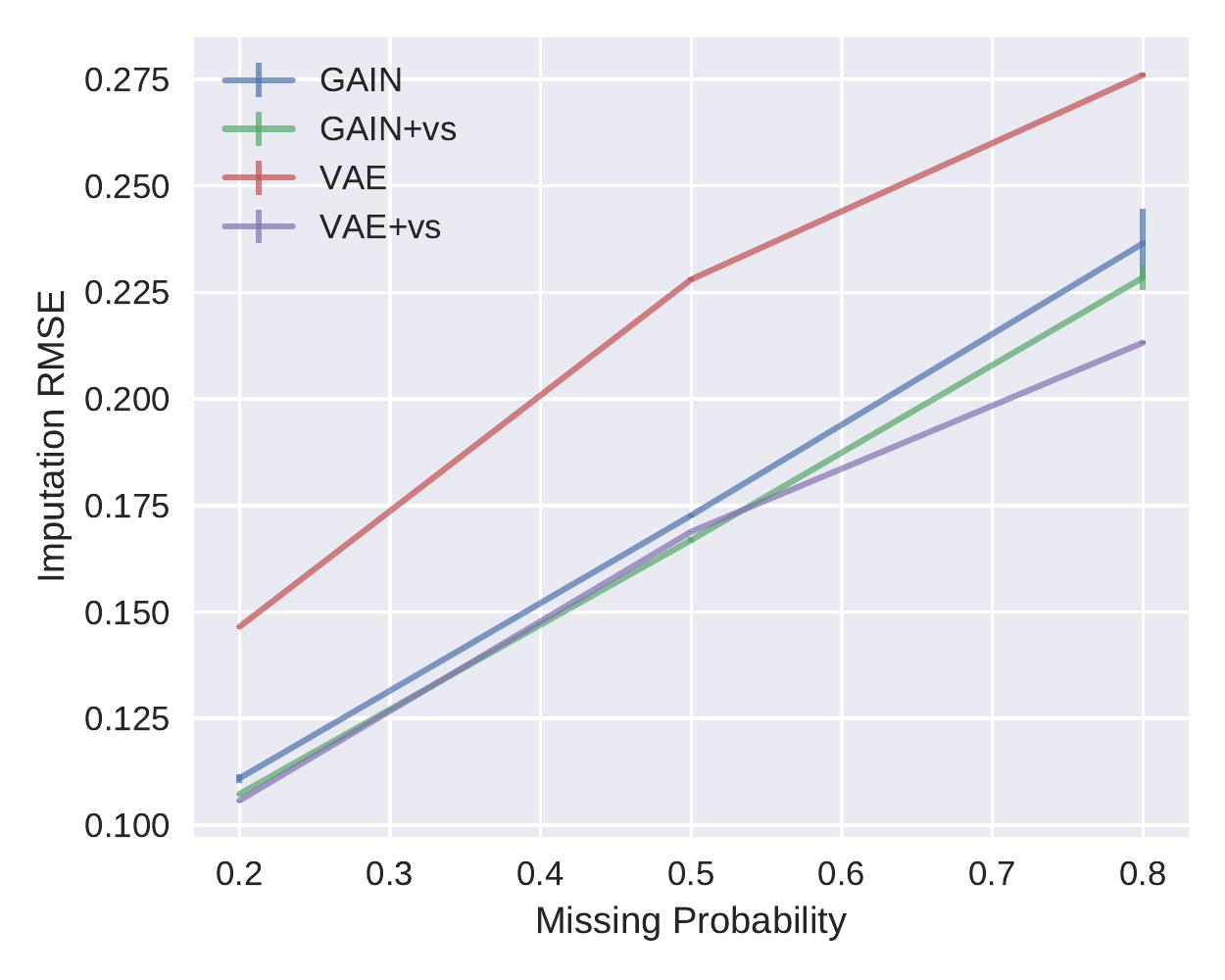}
\endminipage
\caption{Imputation RMSE by increasing proportion of missing values for the Default Credit Card and Online News Popularity datasets.}
\label{fig:categorical}
\end{figure*}

\begin{table}[t]
\caption{Imputation for datasets with numerical variables only.}
\label{tab:imputation-numerical}
\vskip 0.15in
\begin{center}
\begin{scriptsize}
\begin{sc}
\begin{tabular}{lccc}
\toprule
\multicolumn{4}{c}{Breast Cancer} \\
\midrule
Missing & 20\% & 50\% & 80\% \\
\midrule
GAIN & \leavevmode\color{red} $0.049 \pm 0.003$ & \leavevmode\color{red} $0.089 \pm 0.004$ & $0.134 \pm 0.003$ \\
VAE & $0.067 \pm 0.000$ & $0.105 \pm 0.000$ & $0.131 \pm 0.000$ \\
VAE+it & $0.066 \pm 0.000$ & $0.104 \pm 0.000$ & $0.131 \pm 0.000$ \\
VAE+bp & $0.065 \pm 0.000$ & $0.104 \pm 0.000$ & \leavevmode\color{red} $0.129 \pm 0.000$ \\
\midrule
\multicolumn{4}{c}{Letter Recognition} \\
\midrule
Missing & 20\% & 50\% & 80\% \\
\midrule
GAIN & \leavevmode\color{red} $0.069 \pm 0.001$ & $0.110 \pm 0.001$ & $0.145 \pm 0.004$ \\
VAE & $0.070 \pm 0.000$ & \leavevmode\color{red} $0.109 \pm 0.000$ & \leavevmode\color{red} $0.137 \pm 0.000$ \\
VAE+it & $0.070 \pm 0.000$ & \leavevmode\color{red} $0.109 \pm 0.000$ & \leavevmode\color{red} $0.137 \pm 0.000$ \\
VAE+bp & $0.070 \pm 0.000$ & \leavevmode\color{red} $0.109 \pm 0.000$ & \leavevmode\color{red} $0.137 \pm 0.000$ \\
\midrule
\multicolumn{4}{c}{Spambase} \\
\midrule
Missing & 20\% & 50\% & 80\% \\
\midrule
GAIN & $0.024 \pm 0.001$ & $0.035 \pm 0.001$ & $0.045 \pm 0.001$ \\
VAE & \leavevmode\color{red} $0.022 \pm 0.000$ & \leavevmode\color{red} $0.034 \pm 0.000$ & \leavevmode\color{red} $0.045 \pm 0.000$ \\
VAE+it & \leavevmode\color{red} $0.022 \pm 0.000$ & \leavevmode\color{red} $0.034 \pm 0.000$ & \leavevmode\color{red} $0.045 \pm 0.000$ \\
VAE+bp & \leavevmode\color{red} $0.022 \pm 0.000$ & \leavevmode\color{red} $0.034 \pm 0.000$ & \leavevmode\color{red} $0.045 \pm 0.000$ \\
\bottomrule
\end{tabular}
\end{sc}
\end{scriptsize}
\end{center}
\vskip -0.1in
\end{table}

\begin{table}[t]
\caption{Imputation for datasets with mixed variable types.}
\label{tab:imputation-categorical}
\vskip 0.15in
\begin{center}
\begin{scriptsize}
\begin{sc}
\begin{tabular}{lccc}
\toprule
\multicolumn{4}{c}{Default Credit Card} \\
\midrule
Missing & 20\% & 50\% & 80\% \\
\midrule
GAIN & $0.123 \pm 0.001$ & $0.194 \pm 0.002$ & $0.270 \pm 0.005$ \\
GAIN+vs & \leavevmode\color{red} $0.119 \pm 0.004$ & \leavevmode\color{red} $0.184 \pm 0.004$ & $0.255 \pm 0.003$ \\
VAE & $0.176 \pm 0.000$ & $0.275 \pm 0.000$ & $0.339 \pm 0.000$ \\
VAE+it & $0.176 \pm 0.000$ & $0.275 \pm 0.000$ & $0.339 \pm 0.000$ \\
VAE+bp & $0.176 \pm 0.000$ & $0.275 \pm 0.000$ & $0.339 \pm 0.000$ \\
VAE+vs & $0.131 \pm 0.000$ & $0.201 \pm 0.000$ & \leavevmode\color{red} $0.252 \pm 0.000$ \\
VAE+vs+it & $0.131 \pm 0.000$ & $0.201 \pm 0.000$ & \leavevmode\color{red} $0.252 \pm 0.000$ \\
VAE+vs+bp & $0.131 \pm 0.000$ & $0.201 \pm 0.000$ & \leavevmode\color{red} $0.252 \pm 0.000$ \\
\midrule
\multicolumn{4}{c}{Online News Popularity} \\
\midrule
Missing & 20\% & 50\% & 80\% \\
\midrule
GAIN & $0.111 \pm 0.001$ & $0.173 \pm 0.000$ & $0.237 \pm 0.008$ \\
GAIN+vs & $0.107 \pm 0.000$ & \leavevmode\color{red} $0.167 \pm 0.001$ & $0.228 \pm 0.003$ \\
VAE & $0.146 \pm 0.000$ & $0.228 \pm 0.000$ & $0.276 \pm 0.000$ \\
VAE+it & $0.146 \pm 0.000$ & $0.228 \pm 0.000$ & $0.276 \pm 0.000$ \\
VAE+bp & $0.146 \pm 0.000$ & $0.228 \pm 0.000$ & $0.276 \pm 0.000$ \\
VAE+vs & \leavevmode\color{red} $0.106 \pm 0.000$ & $0.169 \pm 0.000$ & \leavevmode\color{red} $0.213 \pm 0.000$ \\
VAE+vs+it & \leavevmode\color{red} $0.106 \pm 0.000$ & $0.169 \pm 0.000$ & \leavevmode\color{red} $0.213 \pm 0.000$ \\
VAE+vs+bp & \leavevmode\color{red} $0.106 \pm 0.000$ & $0.169 \pm 0.000$ & \leavevmode\color{red} $0.213 \pm 0.000$ \\
\bottomrule
\end{tabular}
\end{sc}
\end{scriptsize}
\end{center}
\vskip -0.1in
\end{table}

% references
For the following experiments, we use the notation ``+vs" for variable split, ``+it" for iterative imputation and ``+bp" for backpropagation iterative imputation.
% figures
In Figure \ref{fig:numerical} we present the imputation RMSE by increasing proportion of missing values for datasets with only numerical variables, and in Figure \ref{fig:categorical} the rest of the datasets that contain both numerical and categorical variables.
% iterative algorithms removed from figure
For the sake of clarity, we removed the iterative imputation methods from Figure \ref{fig:categorical}.
% tables
Nevertheless, a full description of the imputation results can be seen on Tables \ref{tab:imputation-numerical} and \ref{tab:imputation-categorical}. 
% observations
From the plots and the tables we can extract the following observations:

\begin{itemize}
    % Decrease in performance
    \item As expected, the performance of every model decreases when the proportion of missing values increases.
    
    % VAE variants are not worth the effort
    \item The iterative and backpropagation alternatives for the VAE imputation do not seem to add improvements over the plain VAE imputation.
    
    % VAE and GAIN are pretty close on the numerical-only datasets
    \item For the datasets with only numerical variables, GAIN seems to be better for the smaller dataset, but it is slightly worse than VAE for the other bigger datasets.

    % The splitting seems to help
    \item The variable splitting appears improve both models, but the improvement is more drastic in the case of VAE.

    % VAE is a lot more stable than GAIN
    \item GAIN has some perceptible variance when trained with different seeds, but all VAE methods seems to be more stable, even when they have worse performance.
\end{itemize}

\subsection{Discussion}

Deep generative models for missing data imputation proved to surpass state-of-the-art methods in previous studies \cite{yoon2018gain,nazabal2018handling}.
However, the power these models offer come at some cost.
The number of hyperparameters to tune is usually larger than traditional non deep learning solutions.
The training time or memory size required for the hyperparameter search using large datasets can be prohibitive in some cases.
Furthermore, the amount of research papers on this domain is growing rapidly, but in many cases, the matching code is either not available or incomplete.
In comparison, traditional methods are well established in the scientific community.
This can lead practitioners to rely on known and robust libraries instead of implementing deep learning alternatives themselves.
\section{Conclusion}

% summarize main idea, state hypothesis, thesis or main research questions
In this paper, we compare several deep generative models for missing data imputation on tabular data, and we propose improvements on top of each model to further improve their imputation quality:
first, using variable splitting to account for categorical variables, and second, an iterative backpropagation-based method for VAEs.

% main findings
Our experiments with public datasets show that our proposals match and improve the performance of deep generative models, which already were shown to outperform state-of-the-art imputation methods in the literature.

% meaning findings, discuss findings, what we learned from the findings
Adding variable splitting techniques to separate variables, applied both on the inputs and the outputs, enhances the imputation power of GAIN and all the presented variants of VAE.
In contrast, both of the VAE iterative imputation procedures did not significantly increase the quality of imputations compared with the basic VAE model.

% suggestions and concluding statement
To support reproducible research, we provide open implementations for our research by
making our code public \footnote{Link blinded for review.}.
We are looking forward to collaborate both with developers and the scientific community.

% \clearpage  % stop the floats here
% \newpage  % the acknowledgments and bibliography go in a new page

\subsubsection*{Acknowledgments}

The experiments presented in this paper were carried out using the HPC facilities of the University of Luxembourg~\cite{VBCG_HPCS14} {\small -- see \url{https://hpc.uni.lu}}.  % comment for review!!!

\bibliography{bibliography}
\bibliographystyle{icml2019}

\end{document}